\documentclass[10pt, letter,twocolumn]{IEEEtran}
\usepackage[dvips]{color}
\usepackage{epsf}
\usepackage{times}
\usepackage{epsfig}
\usepackage{graphicx}
\usepackage{epstopdf}
\usepackage{algorithm}
\usepackage{algorithmic}
\usepackage{amsmath}
\usepackage{amssymb}
\usepackage{amsxtra}
\usepackage{multirow}
\usepackage{mathtools}
\usepackage[font=normalsize]{caption}
\usepackage{cite}	
\usepackage{color}
\usepackage{xcolor}
\usepackage{bbm}
\title{Word Embedding-based Text Processing for Comprehensive Summarization and Distinct Information Extraction}

\author{
\IEEEauthorblockN{\large Xiangpeng Wan, Hakim Ghazzai, and Yehia Massoud}
\IEEEauthorblockA{\small School of Systems and Enterprises, Stevens Institute of Technology, Hoboken, NJ, USA\\ Email: \{xwan6, hghazzai, ymassoud\}@stevens.edu\\ }

{\thanks {\hrule
				\vspace{0.1cm} This paper is accepted for publication in IEEE Technology Engineering Management Society International Conference (TEMSCON'20), Metro Detroit, Michigan (USA).
                
                2020 IEEE. Personal use of this material is permitted. Permission from
                IEEE must be obtained for all other uses, in any current or future media,including reprinting/republishing this material for advertising or promotional purposes, creating new collective works, for resale or redistribution to servers or lists, or reuse of any copyrighted component of this work in other works.
		}}

}

\begin{document}
\maketitle
\begin{abstract}
\boldmath
In this paper, we propose two automated text processing frameworks specifically designed to analyze online reviews. The objective of the first framework is to summarize the reviews dataset by extracting essential sentence. This is performed by converting sentences into numerical vectors and clustering them using a community detection algorithm based on their similarity levels. Afterwards, a correlation score is measured for each sentence to determine its importance level in each cluster and assign it as a tag for that community. The second framework is based on a question-answering neural network model trained to extract answers to multiple different questions. The collected answers are effectively clustered to find multiple distinct answers to a single question that might be asked by a customer. The proposed frameworks are shown to be more comprehensive than existing reviews processing solutions.
\end{abstract}

\section{Introduction}
In modern life, people are more likely to trust their peers over advertising when it comes to purchasing decisions and service selections. 
In fact, according to the Global Trust in Advertising report, which surveyed more than 28,000 Internet respondents in 56 countries, 92$\%$ of customers reveal that they trust recommendations from their friends and relatives above all other forms of advertising, while 70$\%$ of customers trust reviews from other users more than advertising \cite{WordofMouth2012}. Another indication that service/product reviews play an integral role in purchase-decision making process is that two-thirds of US internet users check other online customers' reviews before choosing an article~\cite{LookatReview2018}. 

However, in many cases, hundreds or thousands of reviews exist online for a single product or service and it is impossible for customers to read and check them all. Therefore, it is very worthwhile to provide an efficient review analyzer to process, filter, classify, and extract essential information summarizing the reviews. Natural Language Processing (NLP) is a new emerging AI technology that is used to process and understand textual data for various application such as predicting customers' feelings towards a certain service or product~\cite{liu2015sentiment} or detecting rumors/wrong information on social networks~\cite{7998415}. However, NLP has its limit as it may output inaccurate results due to the fact that machines cannot understand contextual meaning of a review. Another potential solution is the statistical topic modeling approach that aims at discovering the abstract ``topic'' that occur in a collection of documents~\cite{deerwester1990indexing,hofmann2001unsupervised,blei2003latent}. In the paper's context, by extracting the ``topics'' from the reviews about one service/product, the objective is to collect and interpret the ``topics'', e.g., positive points or main issues, that are highlighted by reviewers. However, it is hard to apply such approaches in the context of reviews as the textual input is usually short with low frequency of important words and high number of overlapped and meaningless words. Finally, text summarization methods could also be employed to extract the main bullets outlining long documents~\cite{page1999pagerank,barrios2016variations,robertson2009probabilistic,pal2014approach,vodolazova2013role}. However, it is shown that the performances remain limited and usually, the proposed approaches lack the common terminology and is very linked to the input dataset.
\begin{figure*}[t!]
	\vspace{0.35cm}
	\centerline{\fbox{\includegraphics[width=17cm]{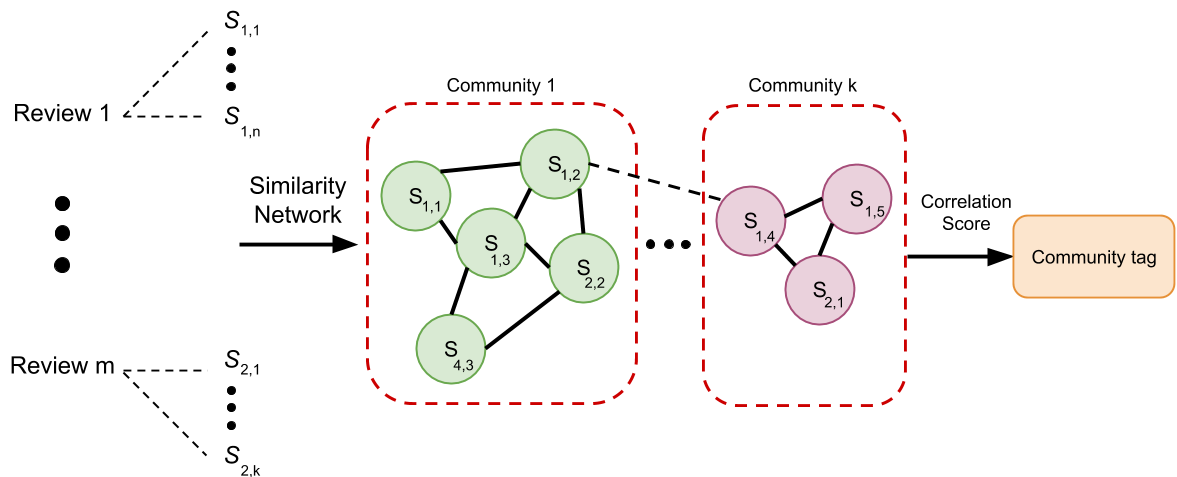}}}
	\caption{\,Framework of review clustering and summarization.   \normalsize}\label{Fig1.1}
\end{figure*}

To cope with the aforementioned limitations of existing approaches and in order to efficiently analyze reviews and extract the most significant information, we propose two generic NLP-based frameworks. The first framework is an unsupervised clustering approach to classify and summarize reviews according to the similarities of the sentences submitted by reviewers such that customers can extract most important feedbacks. It starts by measuring sentences similarities by combining TextRank algorithm, Bidirectional Encoder Representations from Transformers (BERT) model for word-embedding, and a vector dimension matching algorithm. With word-embedding, the words and phrases are converted to a continuous vector space pre-trained on a very large data sets~\cite{mikolov2013distributed} so similarities between two phrases or two sentences can then be calculated by their distances~\cite{goldberg2014word2vec,levy2015improving,dong2017metapath2vec,pennington2014glove}. In 2018, the BERT model, introduced by Google, has redefined the state-of-the-art for eleven NLP tasks~\cite{devlin2018bert} such as text classification, question and answering, and language translation. Afterwards, we apply Louvain method to detect communities of sentences and TextRank algorithm to identify the most meaningful sentences that can tag each community.

Generally, it is not enough to provide customers with a summary about the reviews as usually they may need more specific details about one product/service such as knowing side effect, if any, of a cosmetic product, etc. Therefore, the second framework is designed to extract these kind of details to provide customers with a complete idea about the product/service. Therefore, we adopt a question-answering (QA) model~\cite{kumar2016ask} to rapidly provide answers to a given text from a large volume of reviews. In fact, recently published models, including BERT or XLNet~\cite{zhang2019sg}, enable machines to achieve performance close to human in this challenging area when tested on the Stanford Question Answering Dataset (SQUAD)~\cite{rajpurkar2016squad}. With BERT, we are able to accurately extract details from thousands reviews of service/product using selected questions adopted to the reviews' context. The collected information are then, filtered, clustered, and summarized based on similarity networks to provide customers with decent results. Finally, we apply our proposed frameworks to a practical case of study where we process a google review data about a restaurant in the area of Manhattan, NYC.

\section{Reviews Clustering and Summarization}
\label{Sec2}
In this section, we propose to design a review clustering and summarization model to help customers get an overview/feedback about a product/service by analyzing a large volume of reviews.

\subsection{Methodology}
\begin{figure}[t!]
	\vspace{0.2cm}
	\centerline{\fbox{\includegraphics[width=9cm]{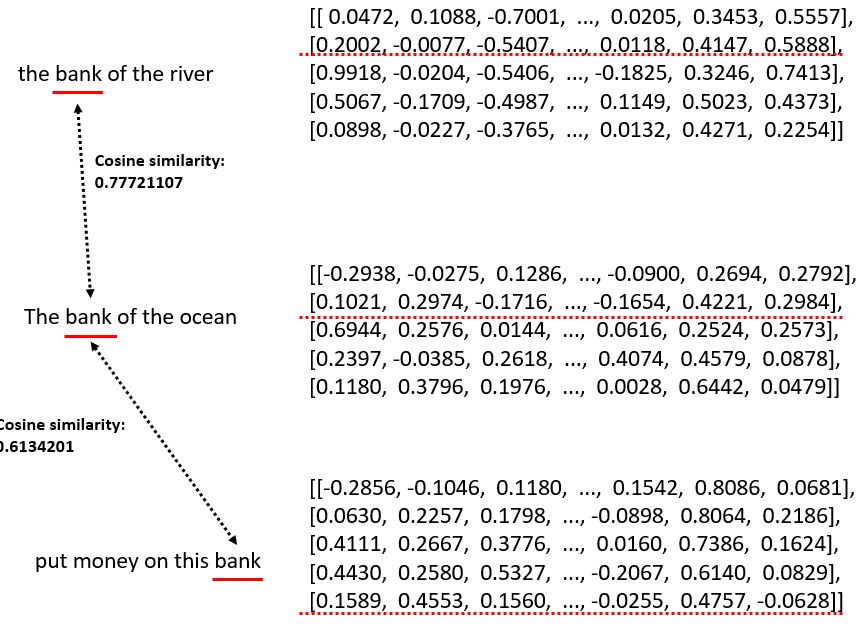}}}\
	\caption{\,Vector space for the same word with different context.   \normalsize}\label{Fig3}
\end{figure}

\begin{figure}[t!]
	\vspace{0.35cm}
	\centerline{\fbox{\includegraphics[width=8.75cm]{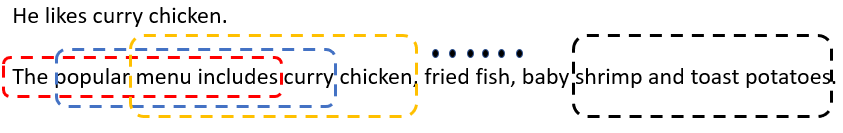}}}
	\caption{\,Procedure to compare between the similarity of two sentences having different lengths.   \normalsize}\label{Fig4}
\end{figure}

The flowchart of the proposed framework for review clustering and summarization is shown in Fig.~\ref{Fig1.1}. It consists of two major parts: the first one creates a similarity network graph joining different sentences collected from the review dataset, while the second part, assigns textual tags for each clustered community. For the first part, the input is constituted by the reviews that we split into independent sentences denoted by $S_{i,j}$ where the $i$ is the review index and $j$ is the index of the sentence in review $i$. Each sentence will be represented by a node in the graph. The edges connecting two nodes represent the similarity between the corresponding sentences. In order to calculate the similarity score denoted by $\sigma_{j,j'}$, we map the sentences to the vector space using word-embedding algorithm and compute the cosine similarity value as the distance between two sentences. Instead of using traditional Word2Vec~\cite{mikolov2013efficient} or GloVe~\cite{pennington2014glove} models, we use BERT to represent each word by a vector having 768 real numbers in $[-1,1]$ as elements. Their values depend on both the context of the sentence and the word itself. So, different vector representations are given for the same word when they are under different contexts. An example is provided in Fig.~\ref{Fig3}. We notice that the word ``bank'' has three different vector representations as the contexts where it is used are different. However, the cosine similarity score is high when the meanings of the word are similar.    
Consequently, if two sentences have $N$ words then, both of them will be represented with a vector space of length $768N$. Hence, the cosine similarity score of the two sentences having the same dimension can be computed. Otherwise, if the other sentence has $M$ words, then a sliding window browsing the longest sentence is applied to compare phrases with the same number of words. Hence, $|N-M|$ comparisons are made and the highest obtained score will represent the similarity between those two sentences as shown in Fig.~\ref{Fig4}, where $\sigma_{1,2} = 0.764$ represents the similarity between sentence $1$ and sentence $2$.

The next step is to create a network graph modeling the similarities between sentences. The vertices of the graphs are the phrases/sentences and the edges connecting two vertices indicate a certain similarity between them. Note that we set a certain threshold for the similarities, the edge exists only if the similarity is larger than the threshold. With the similarity network, our objective is to cluster the sentences into different ``topics'' and assign to them tags by selecting the most meaningful sentences. The clustering is based on the Louvain method designed by Blondel, which is a greedy optimization method that rapidly extract communities from large networks~\cite{blondel2008fast}. In this clustering problem, the objective function to maximize is a modularity metric defined~as follows:
\begin{equation}
\label{c1}
\mathcal{Q} = \frac{1}{2m}\sum_{j,j'} \left[\sigma_{j,j'} - \frac{k_j k_{j'}}{2m}\right]\delta(c_j,c_{j'}),
\end{equation}
where $k_j$ and $k_{j'}$ are the sum of the weights of the edges attached to nodes $j$ and $j'$, respectively, $m$ is the sum of the weights in the graph, $\delta$ is Kronecker delta function with binary value, and $c_j$ and $c_{j'}$ are the communities of the nodes.

\begin{figure}[t!]
	\vspace{0.3cm}
	\centerline{\fbox{\includegraphics[width=8.5cm]{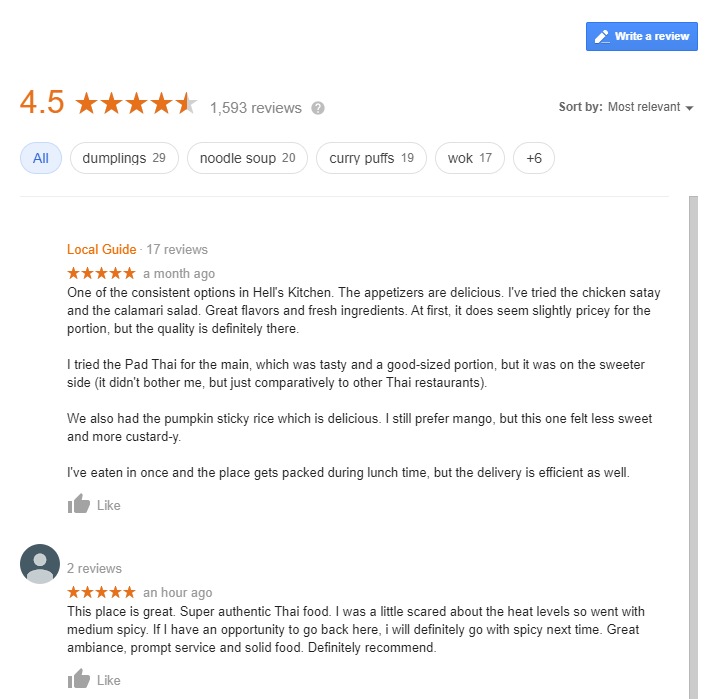}}}
	\caption{\,Example of reviews about the selected restaurant.   \normalsize}\label{Fig2}	\vspace{-0.3cm}
\end{figure}

\begin{figure}[t!]
	\vspace{0.3cm}
	\centerline{\includegraphics[width=9.5cm]{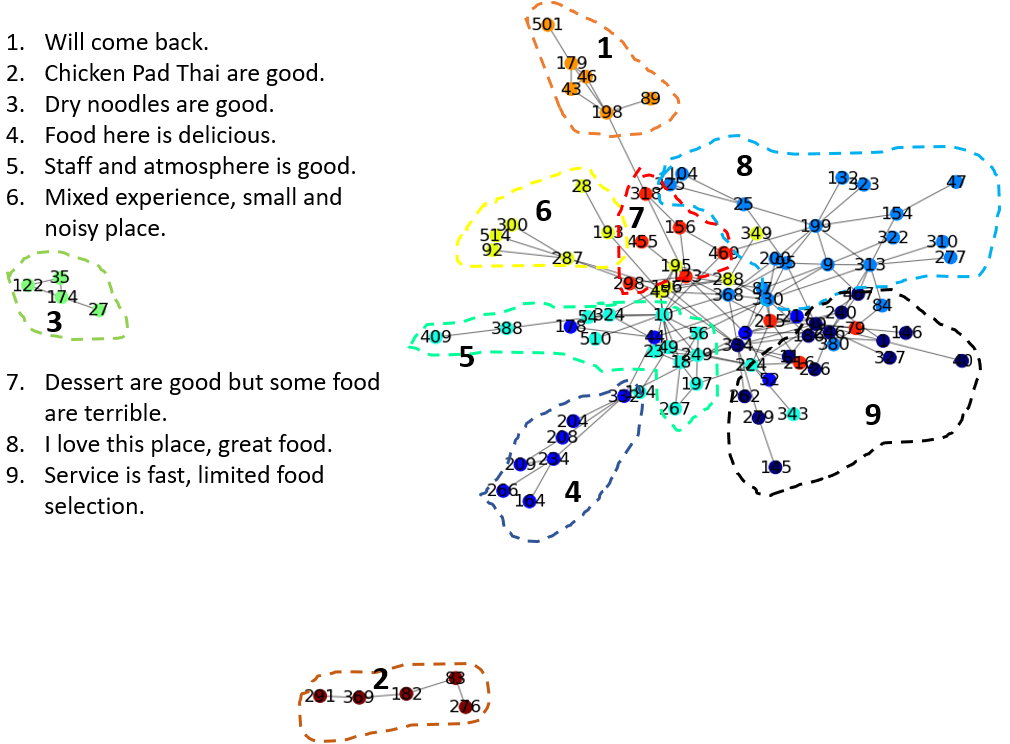}}
	\caption{\,Example of the communities detected from 100 reviews. The sentences having the highest correlation scores are used as tags for each community.   \normalsize}\label{Fig6}	\vspace{-0.3cm}
\end{figure}
Afterwards, we assign another correlation score, denoted by $C_j$, to each sentence in the graph that reflects its similarity with the rest of the phrases (i.e., nodes of the graph) using the TextRank algorithm, which counts the number and quality of links to a sentence to determine how important the corresponding node is. Consequently, the sentences with the highest correlation score is the sentence that is most similar to others and will be used to tag each detected community.

\subsection{Case study: Restaurant Reviews}
In this section, we illustrate and evaluate the output of our proposed framework applied on the case of restaurant reviews. We randomly select a restaurant located in Manhattan, NYC, having 1586 reviews in Google website as shown in Fig.~\ref{Fig2}. From this dataset, we pick the recent one hundred reviews, and feed them into our framework pipeline described earlier. We then split each review into sentences. In case of short phrases, we combine them with previous sentences to avoid having inaccurate results when computing similarities. In Fig~\ref{Fig6}, and for tractability, we provide a graph representing the connections between 568 sentences composing the 100 reviews. The isolated nodes (sentences) are not illustrated. We then highlight the different detected communities colored differently as well as their corresponding tags. Nine independent communities are obtained with the Louvain algorithm, each one is tagged by the sentence having the highest correlation score.  Most of the communities show positive comments since the rating of the restaurant is 4.5, in other words, the positive side of the restaurant is mostly being discussed in the reviews.

\begin{figure*}[t!]
	\vspace{0.3cm}
	\centerline{\fbox{\includegraphics[width=17cm,height=7.5cm]{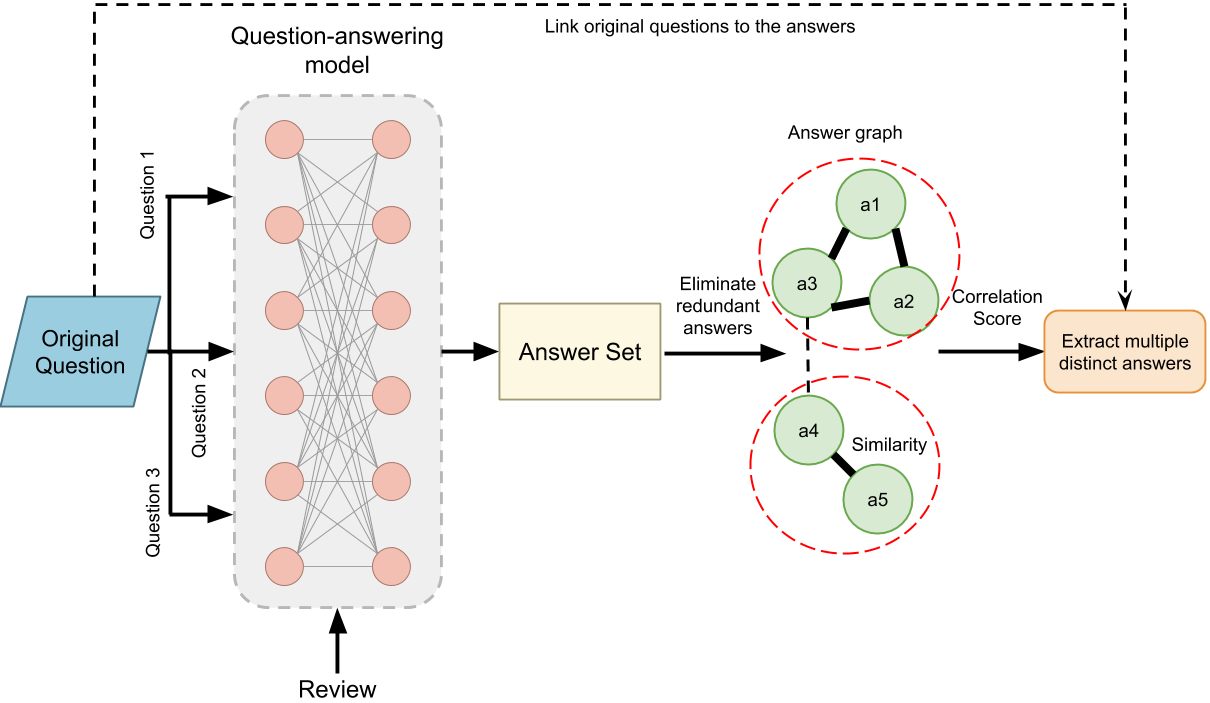}}}
	\caption{\, Proposed framework for multiple distinct answers extraction.   \normalsize}\label{Fig1.2}\vspace{-0.2cm}
\end{figure*}

\section{Multiple Distinct Answers Extraction}
As mentioned in the introduction, having main ideas about a product or service may not be sufficient for certain customers, who also care about details and require deeper information about it. Therefore, in this framework, we propose develop a question-answering model framework to extract details through pre-defined questions that are answered by processing the reviews dataset.

\subsection{Methodology}
To tackle this problem, we propose the second information extraction framework presented in Fig.~\ref{Fig1.2}. It is composed of two major parts: the first part is dedicated to collect answers for the questions that we formulated according to the context of the product/service. We apply the ``BertForQuestionAnswering'' question-answering model trained using the SQUAD dataset~\cite{wolf2019transformers}. The model is batch trained for two epochs using in total 100,000 questions where each batch consists of eight questions. It is shown that the model achieves a matching score of 80.1\% and a F1 score of 83.1\% that are very close to human performance which are 86.8 and 89.5, respectively. Note that the F1 score measures the average overlap between the prediction and ground truth answer\footnote{In our future study, we will explore BERT and ALBERT ensemble models which are expected to achieve better performance over human but require large computational resources.}. The question-answering model provided by BERT is not valid for multi-responses questions. Therefore, to overcome this issue, we proceed by formulating new different questions having the same meaning of the original questions to get all possible answers. The first part outputs, for each question, is a set of answers.

In the second part, we adopt the framework described in Section~\ref{Sec2} so that, for each original question, all the possible answers are collected together and clustered into communities. Then, tags are assigned to them according to their correlation scores. Hence, for each original question, we determine a number of distinct answers corresponding to the number of the detected communities. Optionally, the framework can be used to answer human entered questions by returning the most relevant answers.

\subsection{Case study: Restaurant Reviews}
We employ the same restaurant reviews data from which we pick the most recent one thousand reviews. In Table~\ref{t1}, we present some examples of answers extracted from a one review text by applying the question-answering model. The latter is applied on the first review text (106 words) given in Fig~\ref{Fig2}. 
\begin{table}[h!]
\centering
\caption{Example of the output of the question-answering model applied to the first review given in Fig~\ref{Fig2}}
\begin{tabular}{ |p{3.6cm}||p{3.6cm}|  }
 \hline
 \textbf{Questions} & \textbf{Answers} \\
 \hline
 \hline
 What should I eat?   & ``\textit{The appetizers}''   \\
 \hline
 What can I try? &   ``\textit{I've tried the chicken satay and the calamari salad}''  \\
 \hline
 What is the best food? & ``'' \\
 \hline
 What is delicious?    & ``\textit{pumpkin sticky rice}'' \\
 \hline
 Which dish is recommended? & ``\textit{The appetizers}''  \\
 \hline
 What do you prefer? & ``\textit{mango}'' \\
 \hline
 How is the service? & ``\textit{the delivery is efficient}''  \\
 \hline
 How is the price? &  ``\textit{slightly pricey}''  \\
 \hline
 How long is the waiting time in this place? &  ``.''  \\
 \hline
 Is it clean? &  ``.''  \\
 \hline
\end{tabular}
\label{t1}
\end{table}

From Table~\ref{t1}, we can notice that with the intentional selected questions, we are able to extract most of the required information from a single review including the delicious dish, the comments about the price, and the quality of service in the restaurant. In addition to the original question: ``What is the delicious food to order in the restaurant?'', we use six other similar questions (the first six questions given in the TABLE~\ref{t1}) to extract all the possible answers to the original question. Note that we need to remove the redundant answers, e.g, ``\textit{The appetizers}''. 

\begin{figure}[t!]
	\vspace{0.3cm}
	\centerline{\includegraphics[width=9cm]{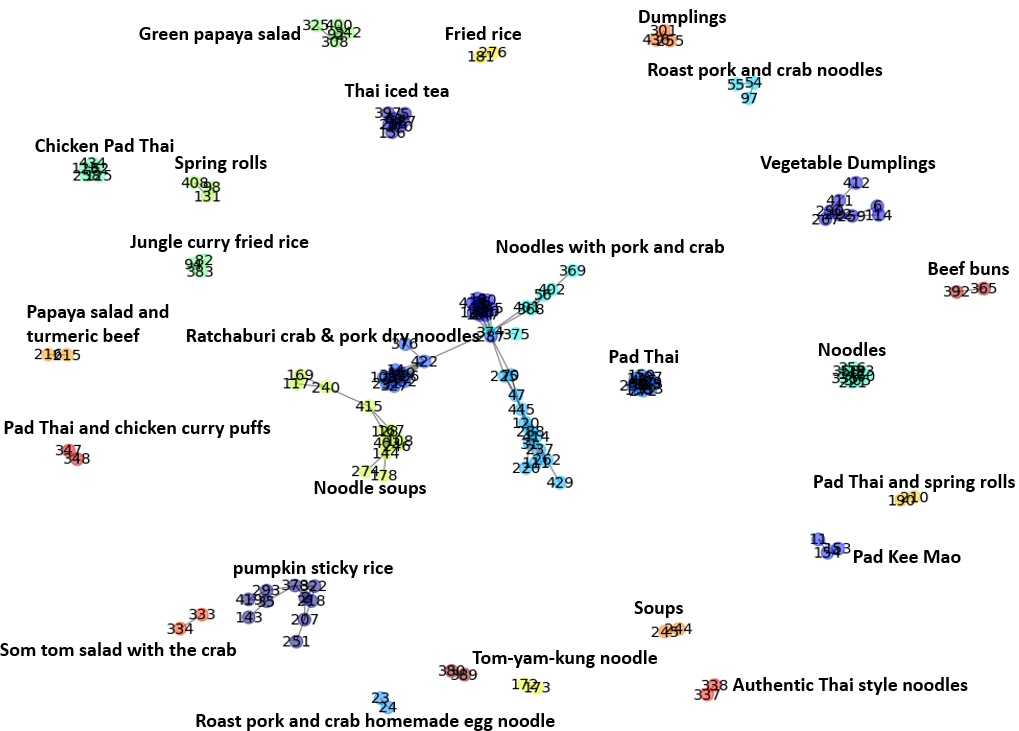}}
	\caption{\,Example of clustered menu items extracted from 1000 reviews.   \normalsize}\label{Fig7}\vspace{-0.2cm}
\end{figure}

In Fig~\ref{Fig7}, we present all the possible answers from one thousand reviews corresponding to this original question. After filtering and clustering the results, we obtain 25 communities with different tags representing 25 different menus items recommended by reviewers. The answer ``\textit{Noodles with pork and crab}'' has the highest correlation score and has been recommended by the highest number of reviewers as it is reflected by the community size.

\begin{figure}[t!]
	\vspace{0.4cm}
	\centerline{\fbox{\includegraphics[width=8cm]{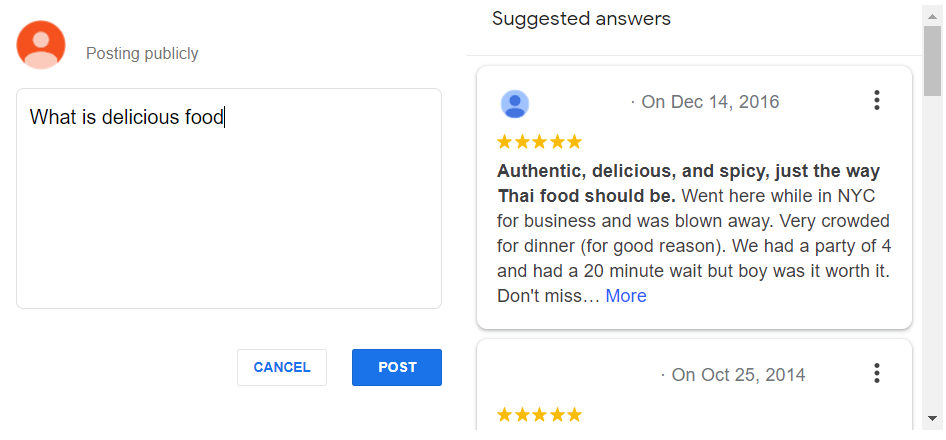}}}
	\caption{\,A snapshot example of the ``ask a question'' service in Google. \normalsize}\label{Fig8}\vspace{-0.2cm}
\end{figure}

Finally, we compare our results with the ``Ask a question'' service provided by Google shown in Fig~\ref{Fig8}. For the same original question, the service provides only ten reviews where seven of them does not provide any useful details. Hence, customers can hardly get comprehensive information or an exhaustive list about their requests. The information provided by the proposed framework are more specific and directed towards the customers need which can ease their purchase decisions.

\section{Conclusion}
\label{Sec4}
In this paper, we proposed two text processing frameworks to provide assistance to customers reviewing previous users' comments. The first framework summarizes the reviews by providing the most important information after clustering their constants in communities and assigning tags to each one of them. The second text processing framework aims to extract detailed information about a product/service by adopting a question-answering neural network model. We also applied the proposed frameworks on a particular case of study and show that our model provides much more comprehensive results than existing solutions.

\bibliographystyle{ieeetr}
\bibliography{main}
\end{document}